\documentclass[sigconf]{acmart}

\usepackage{graphicx}
\graphicspath{ {images/} }
\usepackage{bm}
\usepackage{float}
\usepackage{caption}
\usepackage{fancyhdr}
\usepackage{siunitx}
\usepackage{stfloats}
\usepackage[official]{eurosym}
\usepackage{booktabs}

\copyrightyear{2021}
\acmYear{2021}
\setcopyright{acmcopyright}\acmConference[KDD '21]{Proceedings of the 27th ACM SIGKDD Conference on Knowledge Discovery and Data Mining}{August 14--18, 2021}{Virtual Event, Singapore}
\acmBooktitle{Proceedings of the 27th ACM SIGKDD Conference on Knowledge Discovery and Data Mining (KDD '21), August 14--18, 2021, Virtual Event, Singapore}
\acmPrice{15.00}
\acmDOI{10.1145/3447548.3467090}
\acmISBN{978-1-4503-8332-5/21/08}

\begin{document}
\fancyhead{}

\title{What Happened Next? Using Deep Learning to Value Defensive Actions in Football Event-Data}

\author{Charbel Merhej}
\affiliation{%
  \institution{University of Southampton}
  \city{Southampton}
  \country{United Kingdom}
  }
\email{cm2u19@soton.ac.uk}

\author{Ryan J Beal}
\affiliation{%
  \institution{University of Southampton}
  \city{Southampton}
  \country{United Kingdom}
  }
\email{ryan.beal@soton.ac.uk}

\author{Tim Matthews}
\affiliation{%
  \institution{Sentient Sports}
  \city{Southampton}
  \country{United Kingdom}
  }
\email{tim.matthews@sentientsports.com}

\author{Sarvapali Ramchurn}
\affiliation{%
  \institution{University of Southampton}
  \city{Southampton}
  \country{United Kingdom}
  }
\email{sdr1@soton.ac.uk}

\begin{abstract}
 Objectively quantifying the value of player actions in football (soccer) is a challenging problem. To date, studies in football analytics have mainly focused on the attacking side of the game, while there has been less work on event-driven metrics for valuing defensive actions (e.g., tackles and interceptions). Therefore in this paper, we use deep learning techniques to define a novel metric that values such defensive actions by studying the threat of passages of play that preceded them. By doing so, we are able to value defensive actions based on what they prevented from happening in the game. Our Defensive Action Expected Threat (DAxT) model has been validated using real-world event-data from the 2017/2018 and 2018/2019 English Premier League seasons, and we combine our model outputs with additional features to derive an overall rating of defensive ability for players. Overall, we find that our model is able to predict the impact of defensive actions allowing us to better value defenders using event-data.
\end{abstract}

\ccsdesc[500]{Applied computing}
\ccsdesc[500]{Computing methodologies~Model development and analysis}

\ccsdesc[300]{Computing methodologies~Neural networks}

\keywords{Sports Analytics; Football; Deep Learning; Neural Networks; Applied Machine Learning; Defensive Actions}

\maketitle

\section{Introduction}\label{sect:introduction}

Valuing the actions of humans and agents in the real-world is a problem in many industries. By assigning values to actions committed, we can help evaluate the performance of agents and aid the learning and improvement of future actions. To date, examples of work that explore methods for valuing actions are shown in industrial optimisation \cite{phan2018leveraging}, agent negotiation \cite{liang2012evaluation} and sports analytics \cite{decroos2019actions,schulte2017markov}. Although all these papers aim to assign value to what a human or agent has helped happen, there are many domains in the real-world where some humans/agents are tasked with preventing actions from happening (e.g., in security games agents aim to protect bases or ports and prevent attacks \cite{shieh2012protect}).

One domain where this is also key is in the sports world. In games such as football, American football and basketball, there are players in each team who are tasked with stopping actions (e.g., an American footballer defensive player aims to prevent touchdowns). Therefore in this paper, we focus our attentions on valuing the actions of defenders in Association Football (soccer).\footnote{Referred to as just ``football'' throughout this paper (as it is known outside the U.S.).} 

Due to the rapid advancements in technology in football, the past decade has seen a growing line of data science research and football analysis. Multiple companies\footnote{Examples of these companies include but are not limited to: STATSPerform, WyScout and StatsBomb.} have implemented advanced methods for collecting detailed match actions in the form of event stream data, and more recently in the form of tracking data. The growing availability of these datasets have led to wider applications of artificial intelligence (AI) techniques in football analytics such as machine and deep learning \cite{beal2019artificial}.

An area where this has been particularly true is in the creation of new models for descriptive player performance metrics. This is due to the immense benefits it can provide for football clubs. As explained in \cite{beal2019artificial}, measuring player performance is an important factor in the decision making processes in team sport. However, due to the low scoring and dynamic nature of football, this presents a unique challenge to AI to value actions that often do not impact the scoreline directly but have important longer-term effects \cite{decroos2019actions}. In recent years, multiple successful metrics have been adopted and used universally. One such example is the ``Expected Goals (xG)" metric \cite{xG}, which provides the probability of a shot resulting in a goal in comparison to other similar shots. 

A key limitation of such event-based action models is the lack of application for defensive actions. This is an inherent problem with event-based data due to the fact that defensive actions mainly prevent events from completion and are therefore a challenge to value. This leads to a lot of players not getting truly acknowledged for their defensive duties. To combat this problem, we propose a novel data-driven model that focuses on defensive and out of possession actions by quantifying these actions in a unique and accurate way. What differentiates this model from other traditional football metrics \cite{beal2020optimising, decroos2019actions,xG,altman2015beyond} is that, instead of studying actions that lead to events happening, our model values defensive actions that stop events from occurring. Thus, this paper advances the state of the art in the following ways:

\begin{enumerate}
    \item Introduces a novel model to value defensive actions in football named DAxT (Defensive Action Expected Threat). The model is targeted at the football analytics community and brings together new research in football with deep learning techniques to accurately value defensive actions based on what they have stopped from happening. 
    \item Using real-world data from 760 real-world football games from the past two seasons of the English Premier League (EPL), we train and test our model and find that we can accurately predict the impact of future events to an MAE of 0.015.
    \item We use this model to identify the top football defenders in the EPL and validate this against leading providers of football performance statistics. 
\end{enumerate}

The rest of this paper is structured as follows. In Section \ref{sect:background} we review the literature around player performance metrics in football and introduce basic concepts about defending. Section \ref{sect:modelling} formally defines the goal of our model. In Sections \ref{sect:model} and \ref{sect:experiment} we discuss how we created and solved the model and illustrate the multiple experiments conducted to validate it. Section \ref{sect:modelapplication} shows the results of applying our model on real-world datasets. Finally, in Section \ref{sect:discussion} we discuss the outcomes obtained and we conclude in Section \ref{sect:conclusion}.

\section{Background}\label{sect:background}

In this section, we present an overview of related work regarding player performance metrics. Moreover, we introduce a background on basic concepts in football and the importance of defensive actions. 

\subsection{Related Work}\label{sect:literature}

One of the key areas in football analytics is the study of new player performance metrics. Because of the fast-paced and continuous nature of the game, player performance analysis has been a difficult task in football. As mentioned in \cite{kumar2013machine}, “Baseball, for  example, is a less fluid sports than soccer. It is easier to break down a baseball game into discrete events to be analyzed. The same cannot be said about soccer”. However, with the increased interest in football analytics and advances in technology, more researchers have focused their studies on this specific area. 

\begin{figure}[h]
    \centering
    \includegraphics[width=\linewidth]{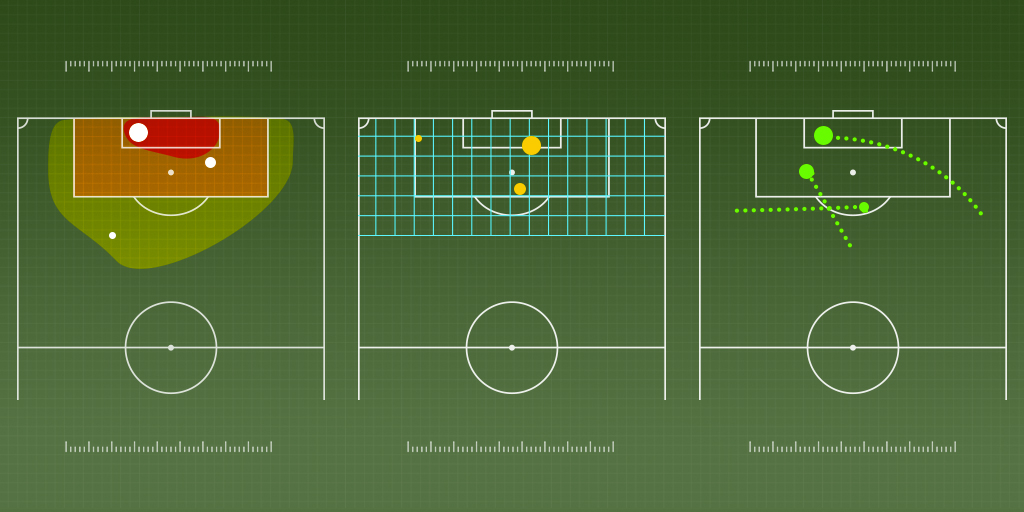}
    \caption{Distance to goal, angle, and type of assist as features used to calculate xG}
    \label{fig:xG}
\end{figure}

As previously mentioned, xG is a metric used to describe the quality of a certain shot. To accomplish this, the authors in \cite{lucey2014quality} analysed the spatio-temporal patterns of the ten-second window of play before a shot for 10,000 shots. In addition to the historical data, they used strategic features such as defender proximity, speed of play, shot location, and game phase to determine the likelihood of a goal being scored. A similar study was done in \cite{xG}, where the authors used distance of the shot taken from goal, the angle in relation to goal and the type of assist as features to calculate xG. The model proved accurate for mid-table teams, whereas the top table teams and lower table teams over and under achieved respectively. Both papers mentioned above used logistic regression in their models. The different features that could be used for xG models are highlighted in Figure \ref{fig:xG} \cite{xGimage}.

Since xG only measures scoring actions, players on the pitch who are not tasked to score goals cannot be evaluated using this metric. This is why the authors in \cite{fernandez2019decomposing} defined a metric called ``Expected Possession Value" (EPV), that quantifies the expected outcome at every moment in a possession. Their model studies passes and turnover probabilities, action likelihood, and pass/ball-drive value expectations that were estimated independently using logistic regression, convolutional neural network and deep neural networks respectively.

Contrarily, instead of measuring periods of possession, the value of each action in a game was studied in \cite{decroos2019actions}. Their VAEP (Valuing Actions by Estimating Probabilities) model calculates this value $V(a_{i})$ by estimating the change in probability of scoring and conceding a goal after every action. This is shown in Equation \ref{eq:vaep}. 

\begin{equation}
    V(a_{i}) = \Delta P_{scores (a_{i},x)} + (-\Delta P_{concedes (a_{i},x)})
    \label{eq:vaep}
\end{equation}

where $\Delta P_{scores (a_{i},x)}$ is the offensive value of team x, denoting the change in probability of scoring after an action $a_{i}$ is committed, and $(-\Delta P_{concedes (a_{i},x)})$ is the defensive value of team x, denoting the change in probability of conceding after $a_{i}$ is committed.

The VAEP metric performed better than traditional player performance metrics and successfully identified promising young players and minor league talents. To estimate the probabilities mentioned above, the authors used features such as the action type, the location of action start and end, time elapsed, etc. Moreover, since the probability of one team scoring is equal to the probability of the other conceding, only two probability estimates were needed for Equation \ref{eq:vaep}. The task was then divided into two separate binary classification problems, where the authors trained both models using CatBoost algorithm.

Another important player performance metric is the Expected Threat (xT) \cite{xT}, which is explained in further details in Section \ref{sect:expectedT}, as it directly correlates to our model.

\subsection{Defending in Football}\label{sect:defending}

Similar to some other team sports, football is an invasion game where the whole team attacks when in possession of the ball, and defends when out of possession. Even though scoring a goal directly contributes to the scoreline, not conceding one is equally as important. Although the main goal of defending is for teams not to concede goals, the way it happens varies according to teams' different tactics and styles of play. Some teams sit deep in their half of the pitch (field) and stack up defenders, while others press high on the field to gain possession back and attack. 

There are two main defensive actions that happen in a football game: Interceptions and Tackles. An interception is when the defending player intentionally intercepts a pass by moving into the line of the intended ball direction\footnote{https://www.mlssoccer.com/glossary}. According to \cite{interceptiondef}, intercepting a pass takes the ability to anticipate, read the play and keep the receiving player guessing. Getting the ball after a misplaced pass is also counted as an interception. 

A tackle can be of 2 different types \cite{tackledef}. A blocking/standing tackle is one where the defender remains on his feet. It is generally used when the player with the ball is coming directly towards the defender. The other type is the sliding tackle, where the defender is off his feet and slides to get the ball \cite{slidingtackle}.

Due to a strong focus in football analysis on attacking metrics that lead to direct contributions, there is a lack of research on the defensive actions mentioned above. In this paper, we aim to quantify different interceptions and tackles happening on the pitch by using our DAxT model - how we model these actions is explained in the following section.

\section{Modelling Defensive Actions}\label{sect:modelling}

In this paper, our goal is to model defensive actions in football to be able to better value the impact they have on the overall game. When modelling attacking actions in games (such as in \cite{decroos2019actions,xG}), the effect of a specific action can be seen according to what happened next and the impact it had on the game. However, the same cannot be said about defensive actions as they usually prevent further actions from happening. Therefore, a logical way to value defensive actions is by assessing what they have prevented from happening and predicting what would have happened if not for that defensive action. By doing so, we present a model for valuing actions that prevent other actions from happening. Although in this paper we focus on football as an application domain, this model could be applied elsewhere where it is important to understand the value of players or agents who prevent actions from happening in games e.g., security games \cite{kiekintveld2009computing,grossklags2008secure} and emergency response \cite{ramchurn2015hac,ramchurn2016human}.

In football, events (such as passes, shots and tackles) happen in a sequence or passage of play. We define a sequence of events as $S = \{e_1, e_2, … , e_N\}$ where $S$ is the sequence, $e$ is an event and $N$ is the number of events in the sequence. Each sequence of events can end in a number of ways, for example the ball may go out of play, there may be a foul leading to a free kick, or a defensive action $DA$ occurs for the opposition to win the ball back from the attacking team. Two key types of defensive actions that we focus on this paper are tackles and interceptions, defined in Section \ref{sect:defending}. These actions can be valued in different ways depending on the game context. 

To value these actions, we aim to predict what was the ‘threat’ of the passage of play that the defensive action has stopped. To calculate this, we use a metric called ‘Expected Threat (xT)’ from \cite{xT}. This model assigns a value to each attacking event based on the probability they have added to the attacking team’s likelihood to score a goal from that passage of play (defined in more details in Section \ref{sect:expectedT}).

We define the function to calculate xT as $f(e)$. We use this model to calculate the xT value of each event $e$ in a sequence $S$, which then allows us to train a machine learning model that is able to predict the xT of $e$ that should have happened after a sequence $S$. This can be defined as $f(e_n)=\Theta(S_{a})$ where $\Theta$ is the trained machine learning model to predict xT, $e_n$ is the event we aim to predict, $S_{a}$ is the passage of play made of $a$ events before a $DA$, and $f(e_n)$ is the xT of the event. 

Using the model to predict the ‘threat’ of an event that did not happen means that we can then value defensive actions by predicting the xT of the event that the defensive action stopped from happening. In the following sections, we give more details regarding the techniques and experiments conducted to create the model described in this paper. 

\section{The DAxT Model}\label{sect:model}

In this section we present how we assign values to the defensive actions discussed in the previous section. We discuss how we applied deep learning to predict the xT of an event that a $DA$ has stopped, which led to the valuation of all interceptions and tackles in games. The first challenge for us to be able to predict what was stopped from happening is assigning the threat of the actions that have happened. This is achieved through the Expected Threat (xT) model discussed in the next subsection.

\subsection{Expected Threat (xT)}\label{sect:expectedT}

The model defined in \cite{xT} aims to measure the threat of any single game state at any moment in a game. In order to do so, the model calculates the probability that the player with the ball will shoot and score, with the probability he will move the ball to another location. This is shown by the following equation:

\begin{equation}
    \textbf{xT}_{x,y}= (s_{x,y} \times g_{x,y}) + (m_{x,y} \times \sum_{z=1}^{16}\sum_{w=1}^{12} T_{(x,y) \rightarrow (z,w)} \textbf{xT}_{z,w}) \label{eq:1}
\end{equation}

The left side of the equation ($s_{x,y} \times g_{x,y}$) represents the probability of shooting from position x,y (how often players opt to shoot from this position), and the probability that this shot is scored (xG \cite{xG} of the shot). For the right side (after the +  sign) , the author is calculating the probability the player will move the ball to another zone, and the value of this movement. It is worth noting that since players can only shoot or move, then $s_{x,y} + m_{x,y} = 100\%$. 

As seen in Equation \ref{eq:1}, in order to calculate the xT at zone (x,y), we need to know beforehand the xT at all other zones. To solve this problem, the author suggests starting with $xT_{x,y} = 0$ at first for all zones, and then evaluating this formula iteratively until convergence. For each evaluation, we use the xT value of the previous iteration. When implementing this model, we ran xT for 43 iterations until full convergence.

\begin{figure}[h]
    \centering
    \includegraphics[width=\linewidth]{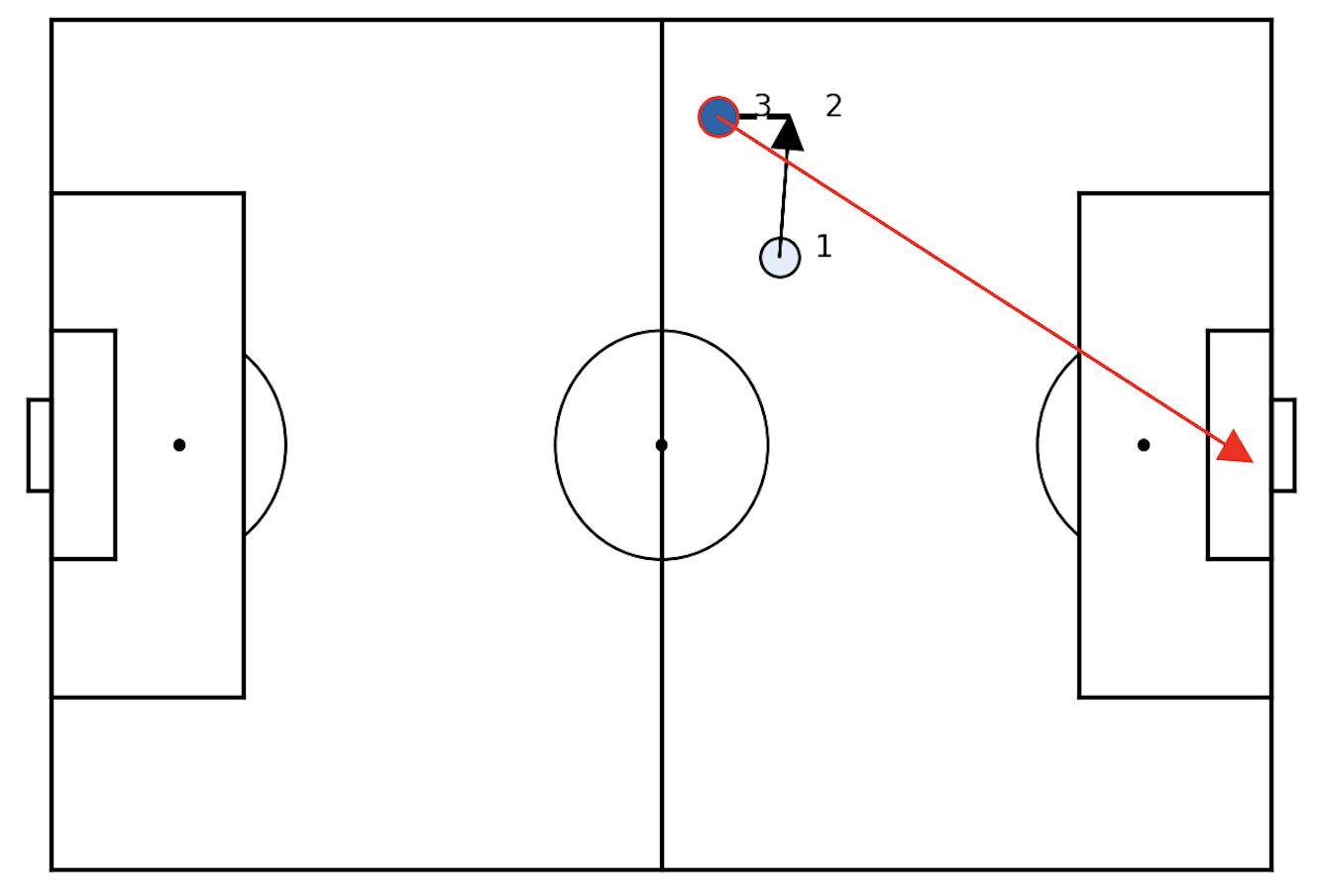}
    \includegraphics[width=\linewidth]{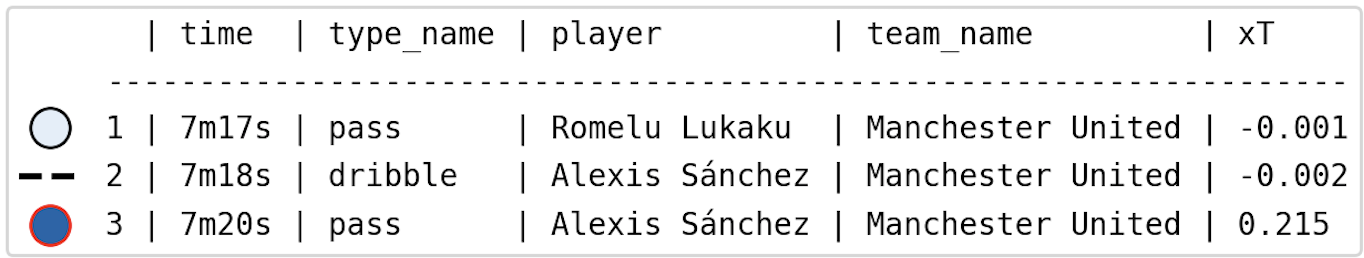}
    \caption{Passage of play showing xT of each action}
    \label{fig:xT}
\end{figure}

To illustrate an example, Figure \ref{fig:xT} shows a passage of play by Manchester United in a game with the xT of each action shown. The first pass (1 to 2) and dribble (2 to 3) have negative xT value since they are moving away from goal, decreasing the probability of the attacking team to score. However, the successful pass of Alexis Sanchez into the penalty box resulted in a large positive xT value, indicating the increased threat of scoring a goal.

In the next subsection we look at how we can use xT to train a model that can predict what would happen after a sequence of events. 

\subsection{Predicting What Was Stopped}\label{sect:predictingwhatstopped}

Due to its generalization mechanism \cite{lawrence1998size}, and application to large sets of data (such as game events), we built a simple neural network model to predict the future xT of the events. Specifically, we have built a Multi Layer Perceptron \cite{tang2015extreme} model using the Keras\footnote{https://keras.io} library. As explained in Section \ref{sect:modelling}, $\Theta$ is fed as input a fixed number of actions $S_{a}$ in a passage of play (Section \ref{sect:exp1} shows the experiment done to determine the number of actions $a$), and as output the xT of the action after this passage of play. Table \ref{tab:input1} shows an example of one input instance $S_{a}$ with $a = 3$. Each event $e$ in $S_{a}$ is represented by its xT ($f(e)$) and x,y coordinates. The output in this case would be ``xT4", signaling the xT of the action after this passage of play.

\begin{table}[h]
    \caption{Input example for our Neural Network model}
    \label{tab:input1}
    \centering
    \begin{tabular}{ | c  c  c  c  c  c  c  c  c |}
    \hline
    xT1 & x1 & y1 & xT2 & x2 & y2 & xT3 & x3 & y3 \\
    \hline
    \end{tabular}
\end{table}

Due to the distribution of the target values being mainly Laplacian with few outliers (explained further in Section \ref{sect:exp3}), the network was trained using a Mean Absolute Error (MAE) loss function, shown in Equation \ref{eq:MAE}.

\begin{equation}
    MAE =  \displaystyle\frac{1}{n}\sum_{t=1}^{n}|\hat{y_t} - y_t| 
    \label{eq:MAE}
\end{equation}

where n is the number of data points used, $\hat{y_t}$ is the predicted xT for a specific instance, and $y_t$ is the actual xT for this instance.

We generate and use three main datasets. The first was formed by collecting all sequences $S_{a}$ (of $a$ consecutive actions) where each sequence precedes a successful event $e$ such that $e \neq DA$. These instances (example in Table \ref{tab:input1}) are used as input for $\Theta$, with the xT of the following event as output. All training and testing (random 80-20 split) of $\Theta$ will be done on this dataset. The other two were made up of valid passages of play $S_{a}$ that came before failed events that were interrupted by a $DA$ (one dataset for interceptions and one for tackles). After validating $\Theta$, it will be fed as input these datasets to predict $f(e_n)$ (equation in Section \ref{sect:modelling}) i.e to predict the xT of event $e_n$ that would have happened without the $DA$. This xT output would thus be the valuation of each interception and tackle.

In the next subsection we discuss how we assign value to each event as well as to the ability of an individual player in a given game for interceptions, tackles, and overall score. 

\subsection{Assigning Value to Events}\label{sect:assignvalue}

As previously mentioned, we apply our DAxT model on the 2 defensive action datasets to predict the xT of the event that was interrupted by a $DA$. After having each interception and tackle assigned its corresponding Interception value $I_{V}$ and Tackle value $T_{V}$, we grouped these defensive actions according to the player committing them. We then calculated the total $I_{V}$ and $T_{V}$ for each player by summing all values together and also computed the average for each per interception and tackle respectively. The results are shown in Section \ref{sect:modelapplication}.

Moreover, we used these metrics to deduce an overall defender score. In total, four features were used to obtain the score: $I_{V}$, $T_{V}$, Clearance xT, and Pass xT. We combined the two metrics we got from the DAxT model with the latter two to derive a final score. Clearance xT is the expected threat of a clearance\footnote{When a player kicks the ball away from their own goal \cite{clearance}.} committed by a defender. The advantage of clearances is that they are moving actions (unlike interceptions and tackles), thus have an xT value which correlates to the DAxT calculated. The final feature is Pass xT, which considers the expected threat of each pass committed by a defender. This feature was used to value the importance of defenders from an attack perspective in build up play, as this is becoming increasingly important at the top levels. 

We used the cumulative values of each feature for each player, and normalized them to a score between 0-100. When simply calculating the average score of the four features, the rankings were topped by defensive-minded players. This is why we introduced weights to the equation. We applied the weights in such a way where the impact of defensive values will be equal to the impact of offensive values. This can be seen in the following equation: 

\begin{equation}
    Sc =  ((I_{V} + T_{V} + CxT)\div 3 + PxT)\div 4
    \label{eq:defenderscore}
\end{equation}

where $Sc$ is the final defender score, $CxT$ is the Clearance xT, and $PxT$ is the Pass xT. The division by 3 was done in order to get the mean of the defensive values and the whole equation was divided by 4 to get the average of all values. Section \ref{sect:defenderscore} illustrates and analyzes the real-world results.

\section{Empirical Evaluation}\label{sect:experiment}

To evaluate and optimise our models, we used a dataset collected from two seasons (2017/18 and 2018/19) from the English Premier League (EPL).\footnote{All data provided by StatsBomb - www.statsbomb.com.} The dataset breaks down each of the games from the tournament into an event-by-event analysis where each event gives different metrics including: event type (e.g., pass, shot, tackle etc.), the pitch coordinates of the event and the event outcome. This is a rich real-world dataset that allows us to rigorously assess the value of our model. The experiments\footnote{Experiments have been run using Keras and TensorFlow.} performed are as follows:

\subsection{Experiment 1: Setting the Model Parameters}\label{sect:exp1}

First, to test the model generalisation, we separated our data into training and validation sets (random split of 80-20). The experiments were then ran on both sets and the MAE (Equation \ref{eq:MAE}) was the metric used.

We ran an experiment to determine the number of actions $a$ that should be taken into consideration in $S_{a}$. There could be an argument about using $a = 1$, since having the xT and location of an event could be enough ``threat" evidence. Another argument would be that using more actions would be beneficial to our model since it is learning more details about the passages of play. Table \ref{tab:numbofactions} shows the training and validation losses, amount of training data, and number of each $DA$ with different number of previous actions.

\begin{table}[h]
    \caption{MAE and instances for different number of actions}
    \label{tab:numbofactions}
    \centering
    \begin{tabular}{| c  c  c  c |}
    \toprule
            & $a = 1$    & $a = 2$    & $a = 3$   \\
    \midrule
    Training Loss  & 0.0174 & \textbf{0.0154} & 0.0123    \\
    Validation Loss & 0.0169 & \textbf{0.0153} & 0.012   \\
    Training Data & 950,139 & \textbf{802,046} & 686,529   \\
    Interceptions & 98,235 & \textbf{75,691} & 59,593   \\
    Tackles & 22,124 & \textbf{17,423} & 13,676 \\
    \bottomrule
    \end{tabular}
\end{table}

The number of training data and defensive actions available decreases as $a$ increases. This is logical, since finding 3 consecutive, successful, and moving actions in the dataset is harder than finding 2 or even 1. Thus, this leads to having less available defensive actions to value as the passages of play become longer.

As highlighted in the table, we chose $a = 2$ as the optimal number of actions for passages of play. This result was selected to balance between minimizing the loss of the model and maximizing the number of defensive actions being valued. Although their number was decreasing, not valuing these $DA$ was worth the increase in accuracy. This is due to the fact that by choosing 2 actions instead of 1, we minimized the randomness of having only $a = 1$ in $S_{a}$ (which could be considered as outliers that confuse the model), thus allowing better and more accurate sequences to be considered. This decrease in $DA$ eliminated the repeated actions that occur in some games, with actions happening quickly after one another that would not interpret a defender's true ability.

\subsection{Experiment 2: Selecting the Features}\label{sect:exp2}

Another experiment was conducted in order to test different feature sets for our model. According to \cite{altman2015beyond}, most existing models that analyze football event data only use location and action type. Since our action type is constant, we tested different combinations of features that include body part, time of game and team ID, other than the already mentioned xT and location. 

After testing different combinations, the results showed that xT and x,y coordinates were truly the most important features. The omission of either one of them (or both) resulted in an increase in MAE > 0.002, whereas trying both features with others resulted in differences of $\approx$ 0.001. The best model according to the lowest MAE value and best learning curve was the model that uses both of these features alone. Adding other variables was either overfitting the model (team ID), which was expected since it disturbs the initial tactical interpretation, or was too general to make a difference (body part, where the large majority of actions was with foot).

\subsection{Experiment 3: Model Prediction}\label{sect:exp3}

The first step towards validating our model is by checking whether its predictive performance deteriorates when applied to unseen data. When exercised on the test set, the model returned an MAE loss of 0.016, which is an accepted result according to the training and validation loss functions in Section \ref{sect:exp1}.\footnote{For comparative purposes a baseline model predicting 0's would have a MAE of 0.0267.} 

The first statistical test we performed was comparing the residuals of the training and testing datasets. The residuals (errors) are the differences between the actual and predicted values of the model. To make sure our model does not over fit to our training data, we used the Levene test \cite{schultz1985levene} and Kolmogorov-Smirnov (KS) \cite{kstest} test on both residuals. These tests are conducted to compare the variance (Levene) and probability distribution (KS) of our training and testing residuals, which are expected to be similar for our model to be considered a good fit. The Levene test returned a statistic value $v = 1.209$ and a p-value $p = 0.272$. Since the p-value is greater than our 5\% threshold, we can then conclude that there is evidence that both residuals have the same variance. Similarly, the KS test's return of $v = 0.0174$ and $p = 0.8333 > 0.05$ suggests that the two probability distributions are the same.

\begin{figure}[h]
    \centering
    \includegraphics[width=\linewidth]{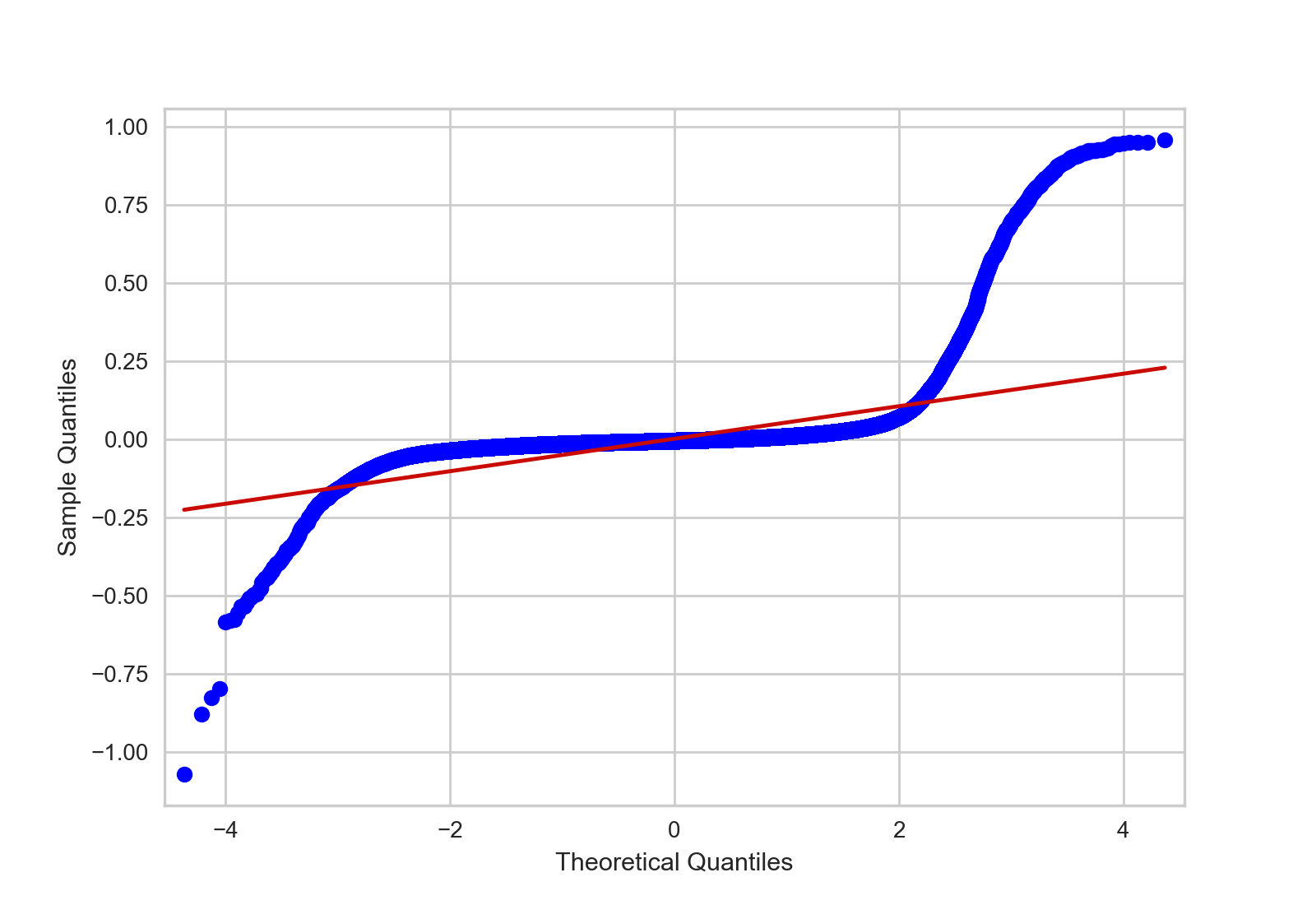}
    \caption{Q-Q plot of normalized residuals}
    \label{fig:residualsnormal}
\end{figure}

When sketching the fitted line of the probability plot (also known as Q-Q plot \cite{kumar2005comparison}) in Figure \ref{fig:residualsnormal}, we could see that the residual's distribution is somewhat normal with long tails on both sides. By visualizing the predictions yielded by our model and comparing them to the actual values, we observed that 96.1\% of the data was between 0.05 and -0.05, explaining the tails in the plot. This reasoning was logical since actions with very high xT (or very low) rarely happen in games compared to regular, less significant actions. 

Finally, to show that there is a correlation between the model's predictions and the actual values, we ran a Pearson correlation test which resulted in an r-value $r = 0.0985$ and a p-value less than 0.05. These results are statistically significant and show that,  using the data available, we have been able to train a model that can predict the xT of the next event in games of football.

\subsection{Experiment 4: Valuing Defenders}\label{sect:exp4}

\begin{figure}[h]
    \centering
    \includegraphics[width=\linewidth]{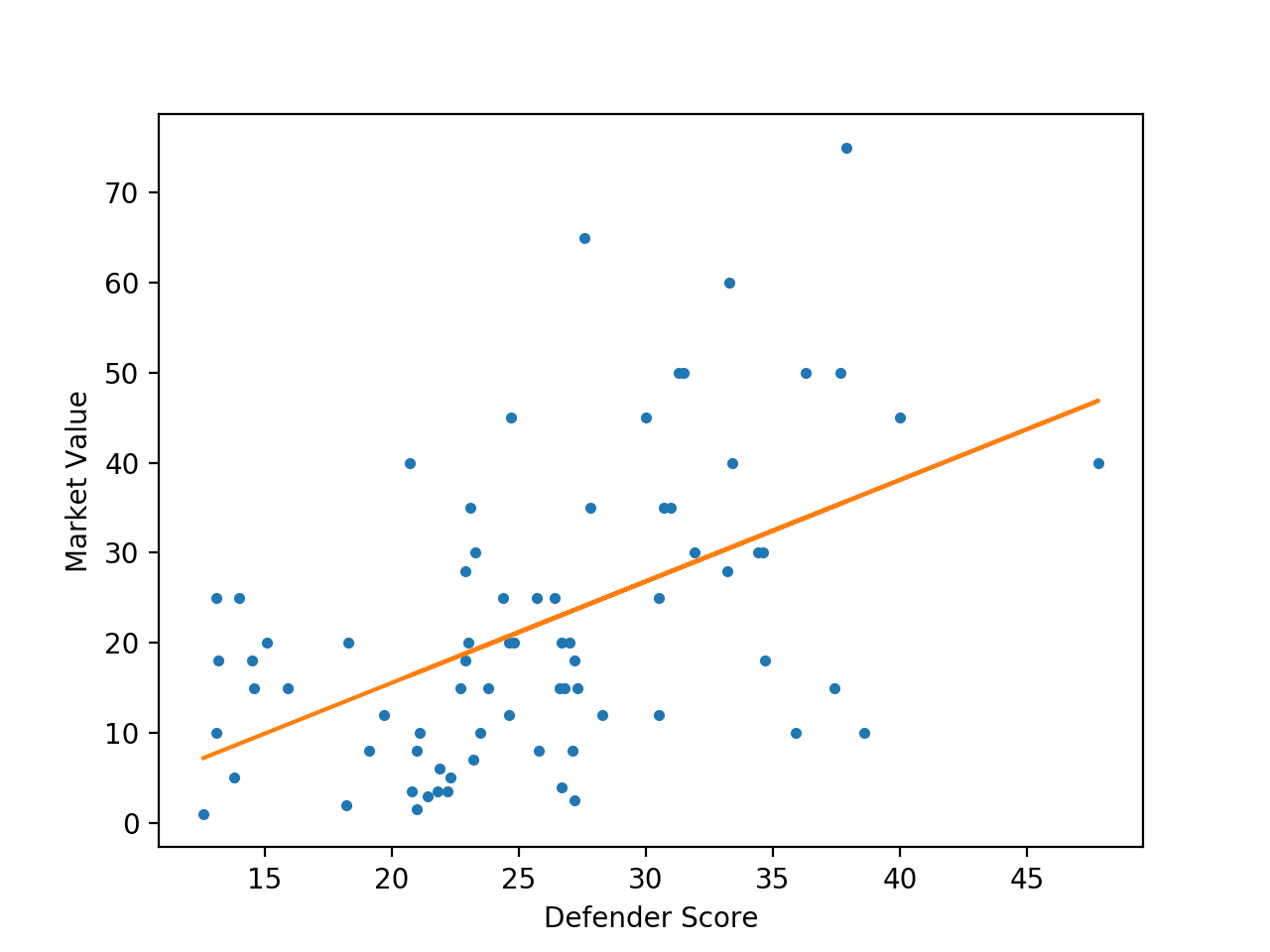}
    \caption{Correlation between Defender Score and Market Value (in millions of \euro{})}
    \label{fig:correlation}
\end{figure}

After applying our model on all defensive actions, we used Equation \ref{eq:defenderscore} to calculate the overall defender score for all players. In order to validate the results, we compared the score of the top 25 center backs, full backs, and defensive midfielders respectively, with their market value\footnote{according to transfermarkt.com} at the beginning of 2019. Figure \ref{fig:correlation} shows the scatter plot that illustrates the correlation between the overall defender score and their market value (in millions of \euro{}).

After plotting the regression line, we calculated the Pearson correlation coefficient $r$ and the p-value. This resulted in $r = 0.533$ and $p = \num{3.37e-6}$. These values show that there is a positive correlation between our defender score and their market value, thus validating the results generated by our model. The reason for not having a stronger correlation coefficient is due to the many factors that add noise to this data away from the actual ability and performance of the player (intangibles that affect market value). For example, other factors include but are not limited to age, nationality, political factors, current club and even popularity \cite{muller2017beyond}. One example in the plot would be Fernandinho who, although considered a world class player and has a high defensive score, was 34 years old in 2019 and arguably past his prime meaning he would have a lower market value.

\section{Model Application to English Premier League}\label{sect:modelapplication}

In this section, we show the results of our model when applied on the English Premier League data for the 2017/18 and 2018/19 seasons. As mentioned in Section \ref{sect:assignvalue}, we separated interceptions from tackles and used our DAxT metric to calculate the Interception Value $I_{V}$ and Tackle Value $T_{V}$. Tables \ref{tab:InterceptionSum} and \ref{tab:TackleSum} show the results after combining all interceptions and tackles respectively for individual players over the 2 Premier League seasons.

The results shown in both tables are logical since almost all players that appear are defenders that play with teams that finished in the bottom half of the table, meaning they tend to commit more defensive actions as their teams on average have a smaller share of possession in each game. This could be normalized based on the possession of the team to assess how much defending a player does. This is also proved in the number of goals conceded column (GC). Although not directly related, committing more defensive actions increases the overall DAxT of defenders as the sum shown in Tables \ref{tab:InterceptionSum} and \ref{tab:TackleSum} are cumulative sums for 2 Premier League seasons.

\begin{table}[h]
    \caption{Top 10 players with most cumulative $I_{V}$}
    \label{tab:InterceptionSum}
    \footnotesize
    \centering
    \begin{tabular}{| c  c  c  c  c |}
    \toprule
     \textbf{Player Name}    &  \bm{$I_{V}$}    & \textbf{Count}  & \textbf{GC} & \textbf{Team} \\
    \midrule
    James Tarkowski   & 1.605 & 641 & 93 & Burnley \\
    Ben Mee & 1.576 & 660  & 95 & Burnley \\
    Nathan Aké & 1.537 & 616 & 129 & AFC Bournemouth\\
    Lewis Dunk & 1.508 & 566 & 109 & Brighton \& Hove Albion \\
    Shane Duffy & 1.473 & 604 & 106 & Brighton \& Hove Albion \\
    Harry Maguire & 1.457 & 593  & 100 & Leicester City \\
    Virgil van Dijk & 1.447 & 657 & 50 & Southampton/Liverpool \\
    Christopher Schindler & 1.411 & 620 & 124 & Huddersfield Town \\
    Jamal Lascelles & 1.386 & 565 & 67 & Newcastle United \\
    Steve Cook & 1.357 & 612 & 109 & AFC Bournemouth \\
    \bottomrule
    \end{tabular}
\end{table}

\begin{table}[h]
    \caption{Top 10 players with most cumulative $T_{V}$}
    \label{tab:TackleSum}
    \small
    \centering
    \begin{tabular}{| c  c  c  c  c |}
    \toprule
     \textbf{Player Name}    &  \bm{$T_{V}$}   & \textbf{Count} & \textbf{GC} & \textbf{Team}\\
    \midrule
    Wilfred Ndidi  & 0.435 & 169 & 95 & Leicester City \\
    Christopher Schindler & 0.4 & 118 & 124 & Huddersfield Town \\
    Idrissa Gana Gueye & 0.384 & 142 & 86 & Everton \\
    James Tarkowski & 0.38 & 85 & 93 & Burnley \\
    César Azpilicueta & 0.353 & 120 & 77 & Chelsea \\
    Nathan Aké & 0.335 & 90 & 129 & AFC Bournemouth \\
    Luka Milivojević & 0.318 & 110 & 91 & Crystal Palace \\
    Aaron Wan-Bissaka & 0.315 & 108  & 56 & Crystal Palace \\
    Ben Mee & 0.292 & 68 & 129 & Burnley \\
    Harry Maguire & 0.286& 67 & 100 & Leicester City  \\
    \bottomrule
    \end{tabular}
\end{table}

To get a clearer idea about which players carried out the most important $DA$, we normalized the values by $DA$ committed i.e. showing the average $I_{V}$ and $T_{V}$ for all interceptions and tackles committed respectively by each player. We set thresholds to only include players that committed more than 100 interceptions and more than 50 tackles (this is so we remove the players who had not played a high number of minutes and skew the results). Tables \ref{tab:InterceptionAverage} and \ref{tab:TackleAverage} show the top 10 players with highest average values for each category.

\begin{table}[h]
    \caption{Top 10 players with highest average $I_{V}$ per interception}
    \label{tab:InterceptionAverage}
    \scriptsize
    \centering
    \begin{tabular}{| c  c  c  c  c |}
    \toprule
     \textbf{Player Name}    &  \bm{$I_{V}$} \textbf{Average}    & \textbf{Count}  & \textbf{GC} & \textbf{Team}\\
    \midrule
    Fabián Balbuena   & 0.0031 & 232 & 32 & West Ham United \\
    James Ward-Prowse & 0.00306 & 125  & 92 & Southampton \\
    Paul Pogba & 0.00305 & 138 & 70 & Manchester United \\
    Cheikhou Kouyaté & 0.00294 & 176 & 97 & West Ham United/Crystal Palace  \\
    Geoff Cameron & 0.00294 & 122 & 30 & Stoke City\\
    Isaac Hayden & 0.00291 & 162 & 65 & Newcastle United \\
    Jefferson Lerma & 0.0029 & 140 & 61 & Bournemouth \\
    Solomon March & 0.00288 & 125 & 106 & Brighton \& Hove Albion \\
    Patrick van Aanholt & 0.00288 & 289 & 84 & Crystal Palace  \\
    Joe Bennett & 0.00287 & 168 & 50 & Cardiff City \\
    \bottomrule
    \end{tabular}
\end{table}

\begin{table}[h]
    \caption{Top 10 players with highest average $T_{V}$ per tackle}
    \label{tab:TackleAverage}
    \footnotesize
    \centering
    \begin{tabular}{| c  c  c  c  c |}
    \toprule
     \textbf{Player Name}    &  \bm{$T_{V}$} \textbf{Average}    & \textbf{Count} & \textbf{GC} & \textbf{Team} \\
    \midrule
    James Tarkowski & 0.00448 & 85  & 93 & Burnley \\
    Shane Duffy & 0.00444 & 54 & 106 & Brighton \& Hove Albion \\
    Jamal Lascelles & 0.00443 & 53  & 67 & Newcastle United\\
    Ben Mee & 0.00429 & 68 & 95 & Burnley \\
    Harry Maguire & 0.00427 & 67 & 100 & Leicester City \\
    James Tomkins & 0.00426 & 62 & 69 & Crystal Palace \\
    Issa Diop & 0.00416 & 51 & 43 & West Ham United \\
    Wesley Hoedt & 0.00414 & 69 & 66 & Southampton\\
    Adrian Mariappa   & 0.00413 & 53 & 74 & Watford \\
    David Luiz & 0.00404 & 51  & 49 & Chelsea/Arsenal \\
    \bottomrule
    \end{tabular}
\end{table}

The inputs were first standardized when training the model, but an inverse transform function was then applied to get back the original corresponding output values. This way, the outputs can be more interpretable and understandable i.e. Fabian Balbuena’s $I_{V}$ average directly correlates to the fact that on average, his interceptions prevented actions of xT = 0.0031 of happening.

\section{Discussion}\label{sect:discussion}

One advantage this model offers is a framework to compare defensive players against each other. To take 2 examples, we used N'Golo Kanté and Aaron Wan-Bissaka as benchmarks \cite{benchmarks}. Although both players are world class in many ways, their most well-known important trait (affects their market value) is their ability to intercept (Kanté) and tackle (Wan-Bissaka). We searched for players having similar number of defensive actions as both and compared their defensive and market values.\footnote{according to transfermarkt.com}

\begin{table}[h]
    \caption{Comparing players by $I_{V}$ and market value}
    \label{tab:compareIV}
    \footnotesize
    \centering
    \begin{tabular}{| c  c  c  c |}
    \toprule
    \textbf{Player Name} & \bm{$I_{V}$} \textbf{Average} & \textbf{Interceptions} & \textbf{Market Value} \\
    \midrule
    N'Golo Kanté  & 0.00255 & 247 & \euro{100M}   \\
    Pierre-Emile Højbjerg & 0.00216 & 241 & \euro{12M}  \\
    Fernando & 0.00205 & 264 & \euro{5M}  \\
    \bottomrule
    \end{tabular}
\end{table}

\begin{table}[h]
    \caption{Comparing players by $T_{V}$ and market value}
    \label{tab:compareTV}
    \small
    \centering
    \begin{tabular}{| c  c  c  c |}
    \toprule
    \textbf{Player Name} & \bm{$T_{V}$} \textbf{Average} & \textbf{Tackles} & \textbf{Market Value} \\
    \midrule
    Aaron Wan-Bissaka  & 0.00291 & 108 & \euro{30M}   \\
    DeAndre Yedlin & 0.00139 & 89 & \euro{8M}  \\
    Pablo Zabaleta & 0.00128 & 97 & \euro{4M}  \\
    \bottomrule
    \end{tabular}
\end{table}

In Tables \ref{tab:compareIV} and \ref{tab:compareTV}, our benchmark players are compared with other players having similar number of defensive actions and playing in the same position. Their $I_{V}$ and $T_{V}$ average per interception and tackle respectively are observed, as well as their market value at the beginning of 2019. As we can see in both tables, the $I_{V}$ and $T_{V}$ are directly proportional to the market value, with our benchmark players having much higher market value than the others. Without the results yielded by our model, we would only be able to compare the raw numbers of interceptions and tackles, without truly understanding why our benchmarks players are top of the class in their defensive ability.

To make sure our model is outputting logical results, we plotted the defensive actions of players in each category and colored each action according to its DAxT. The blue color signals a very high DAxT (top 10 \% of all DAxT in the dataset), with green (top 30\%), yellow (top 50\%) and red (the rest) coming after.

\begin{figure}[h]
    \centering
    \includegraphics[width=\linewidth]{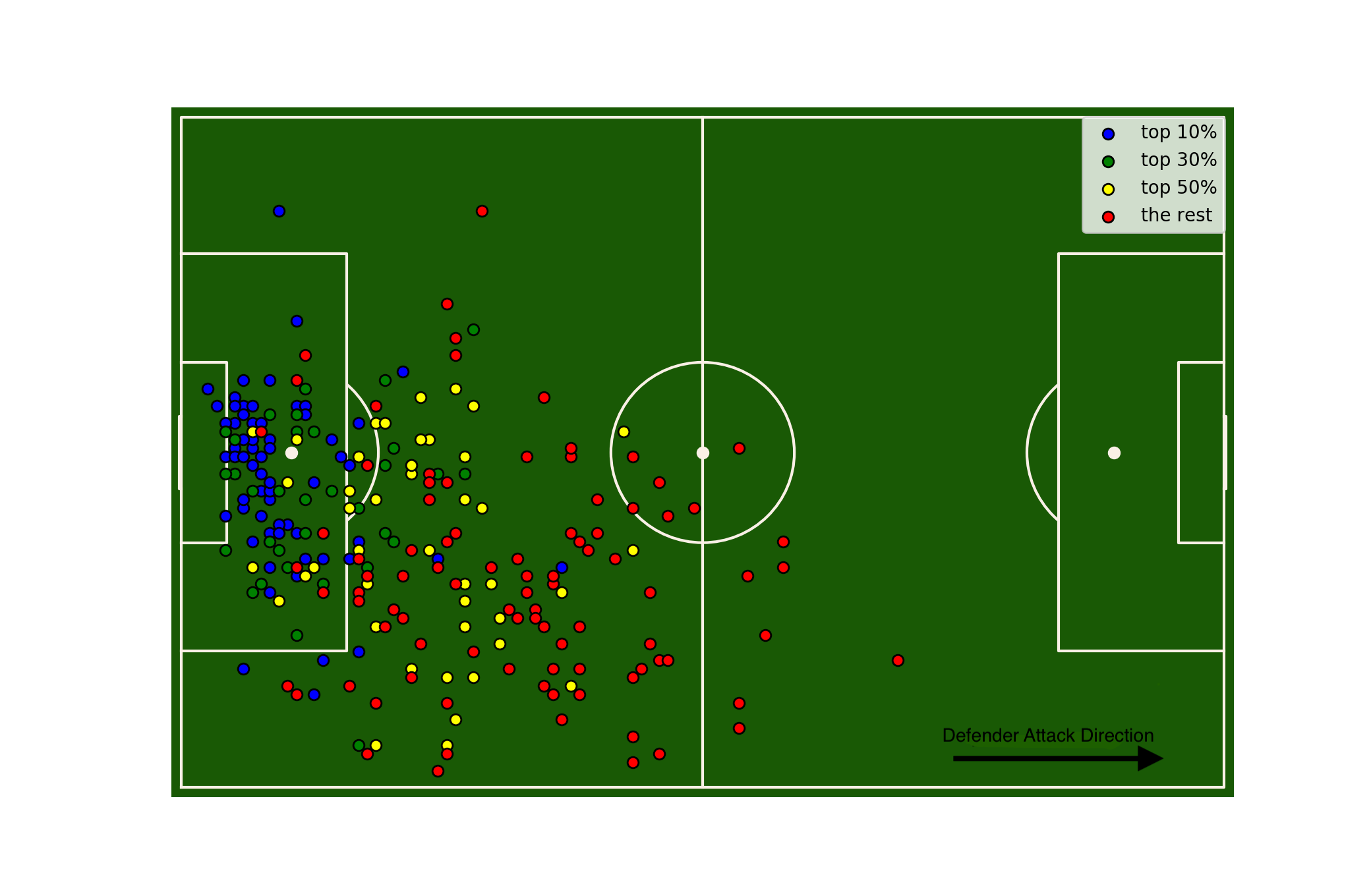}
    \caption{Interceptions committed by Fabian Balbuena}
    \label{fig:balbuena}
\end{figure}

\begin{figure}[h]
    \centering
    \includegraphics[width=\linewidth]{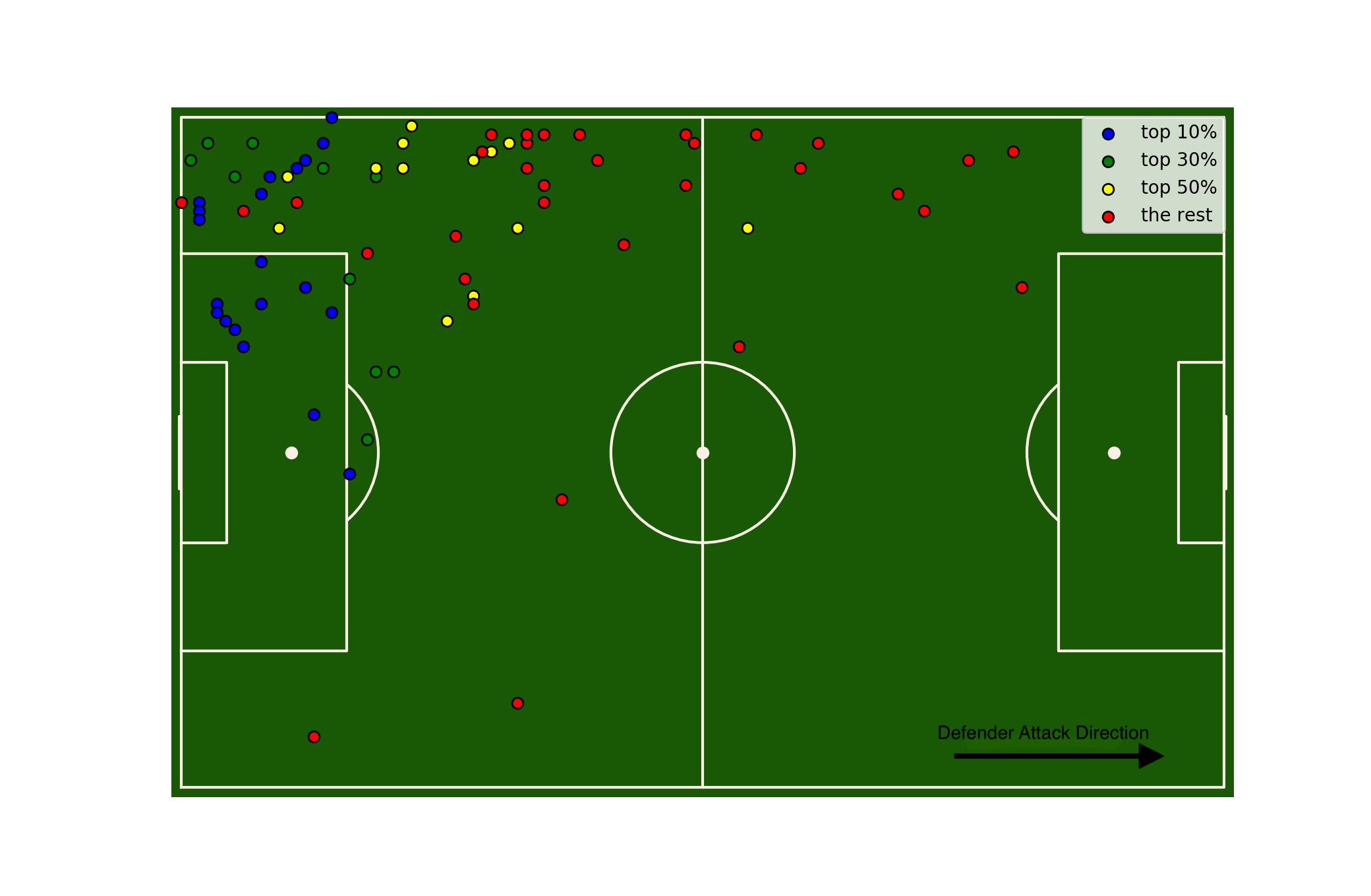}
    \caption{Tackles committed by Andrew Robertson }
    \label{fig:robertson}
\end{figure}

Figure \ref{fig:balbuena} illustrates the interceptions committed by Fabian Balbuena. The ones closest to goal (defensively) are shown to be the most important defensive actions (top 10\%). These interceptions illustrate that the preceding passages of play were of high threat, thus increasing the $I_{V}$ of each interception. The values gradually decrease as actions occur further up the pitch, which is logical since passages of play in the middle of the pitch do not usually lead to conceding goals. The same concept applies to Figure \ref{fig:robertson}, that illustrates the tackles committed by Robertson. The ones which directly led to stopping actions of high threat are in blue, and most of the tackles are on the left side of the pitch due to Robertson's position as a left back.

\subsection{Overall Defender Score}\label{sect:defenderscore}

We applied the overall defender score (introduced in Section \ref{sect:assignvalue}) for ranking and filtered out according to position (center back or full back). We then used CIES Football Observatory and InStat Performance Index \cite{index} for the 2019 season as a means of comparison for our rankings (CIES). Moreover, we calculated the number of goals conceded (GC), appearances (A), goals conceded per appearance (GC/A), and assists (AST) for full backs. Table \ref{tab:CBranking} shows our rankings for center backs while Table \ref{tab:FBranking} shows the rankings for full backs. 

\begin{table}[h]
    \caption{Ranking Center Backs according to defender score}
    \label{tab:CBranking}
    \small
    \centering
    \begin{tabular}{| c  c  c  c  c  c  c |}
    \toprule
    \textbf{Player Name} & \textbf{Score} & \textbf{GC} & \textbf{A} & \textbf{GC/A} &  \textbf{Rank} & \textbf{CIES} \\
    \midrule
    Virgil Van Dijk  & 37.901 & 50 & 64 & 0.781 & 1 & 1 \\
    Harry Maguire &  36.3 & 100 & 69 & 1.449 & 2 & 4\\
    Ben Mee & 35.942 & 95 & 67 & 1.418 & 3 & 37 \\
    James Tarkowski & 34.733 & 93 & 66 & 1.409 & 4 & 13\\
    Shkodran Mustafi & 34.636 & 70 & 58 & 1.207 & 5 & 20\\
    Jan Vertonghen & 34.426 & 54 & 58 & 0.931 & 6 & 20\\
    Toby Alderweireld & 33.459 & 44 & 48 & 0.916 & 7 & 11\\
    Aymeric Laporte & 33.284 & 26 & 44 & 0.591 & 8 & 6\\
    Nathan Aké & 33.25 & 129 & 76 & 1.697 & 9 & 31\\
    Michael Keane & 31.913 & 85 & 63 & 1.349 & 10 & 37\\
    Lewis Dunk & 30.572 & 109 & 74 & 1.472 & 11 & 26 \\
    David Luiz & 30.536 & 49 & 46 & 1.065 & 12 & 20\\
    Antonio Rüdiger & 29.975 & 57 & 60 & 0.95 & 13 & 17\\
    Shane Duffy & 28.295 & 106 & 72 & 1.472 & 14 & 20\\
    Nicolas Otamendi & 27.826 & 32 & 52 & 0.615 & 15 & 17\\
    \bottomrule
    \end{tabular}
\end{table}

\begin{table}[h]
    \caption{Ranking Full Backs according to defender score}
    \label{tab:FBranking}
    \scriptsize
    \centering
    \begin{tabular}{| c  c  c  c  c  c  c  c |}
    \toprule
    \textbf{Player Name} & \textbf{Score} & \textbf{GC} & \textbf{A} & \textbf{GC/A} & \textbf{AST} & \textbf{Rank} & \textbf{CIES} \\
    \midrule
    César Azpilicueta  & 47.823 & 77 & 75 & 1.026 & 11 & 1 & 6\\
    Andrew Robertson &  40.047 & 38 & 58 & 0.655 & 16 & 2 & 2\\
    Patrick van Aanholt & 38.568  & 84 & 64 & 1.3125 & 3 & 3 & 24\\
    Kyle Walker & 31.464  & 39 & 65 & 0.6 & 7 & 4 & 12\\
    Trent Alexander-Arnold & 31.31 & 32 & 48 & 0.66 & 13 & 5 & 5\\
    Kieran Trippier & 31.082 & 44 & 51 & 0.86 & 8 & 6 & N/A\\
    Ryan Bertrand & 27.206 & 91 & 59 & 1.54 & 4 & 7 & 27\\
    Pablo Zabaleta & 26.69 & 97 & 63 & 1.54 & 2 & 8 & N/A\\
    Ben Davies  & 26.44 & 46 & 56 & 0.82 & 6 & 9 & 24\\
    Marcos Alonso & 24.691  & 58 & 64 & 0.906 & 6 & 10 & 1\\
    Ben Chilwell & 24.442  & 73 & 60 & 1.216 & 6 & 11 & 8\\
    Aaron Wan-Bissaka & 23.331  & 56 & 42 & 1.33 & 3 & 12 & 9\\
    Ricardo Pereira  & 23.046 & 45 & 35 & 1.285 & 6 & 13 & 4\\
    Luke Shaw & 22.994 & 41 & 40 & 1.025 & 4 & 14 & 13\\
    Cédric Soares & 22.73 & 75 & 50 & 1.5 & 5 & 15 & 19\\
    \bottomrule
    \end{tabular}
\end{table}

Although there is a season difference between the rankings, we can see many similarities in player standings. Most of the players mentioned are found in both studies, assuring that our results are logical. Moreover, our findings match the top defenders identified by experts in the press for that season\footnote{https://www.mirror.co.uk/sport/football/news/top-10-central-defenders-premier-13772574}. This indicates the importance of applying statistical models such as our metrics and xT, rather than using raw data. 

Furthermore, the tables illustrate consistency between our defender score and the other variables. For example, we can see that 5 of our top 6 full backs have a low average of goals conceded/appearance with a high number of assists relative to the rest. Also, although Harry Maguire and Nathan Aké, for instance, have a high number of goals conceded/appearance due to the teams they were playing for, their transfer fees (transfermarkt.com) of \euro{87M} to Manchester United and \euro{45M} to Manchester City respectively show why they are top of our ranking.

As different teams have different playing styles, valuing defenders as such has many advantages. Other than using it directly to compare between players, scouts and teams could use the same metrics but change the weights according to their needs. Instead of equally weighing the defensive and attacking side of players, some teams (such as relegation-threatened ones) could want to focus more on specific defensive metrics, weighing each one accordingly. Moreover, the same defender score could also be used for defensive midfielders. Table \ref{tab:DMranking} shows the top 10 defensive midfielders according to score. 

\begin{table}[h]
    \caption{Ranking Defensive Midfielders according to defender score}
    \label{tab:DMranking}
    \centering
    \begin{tabular}{| c  c  c |}
    \toprule
    \textbf{Rank} & \textbf{Player Name} & \textbf{Score} \\
    \midrule
    1 & Granit Xhaka  & 37.648  \\
    2 & Fernandinho &  37.437   \\
    3 & Nemanja Matić & 30.737    \\
    4 & Jorginho & 27.652   \\
    5 & Luka Milivojević & 27.299  \\
    6 & Abdoulaye Doucouré & 25.734  \\
    7 & Idrissa Gueye  & 24.633   \\
    8 & Oriol Romeu & 24.629  \\
    9 & Mark Noble  & 23.197  \\
    10 & Wilfred Ndidi & 23.107   \\
    \bottomrule
    \end{tabular}
\end{table}

\subsection{Remaining Challenges}\label{sect:futurework}

One limitation of our DAxT model is that some defensive actions will be valued less than they should be due to unique passages of plays. For instance, the opposition could counterattack and be in a 3vs1 situation (3 players attacking while only 1 is defending) in the middle of the pitch. If the defender intercepts the ball, our model would not highly reward this action because similar passages of play are not usually of high threat (without taking into consideration the number of defenders behind the ball). This is why future work for valuing defenders' contributions should be centered around the growing collection of tracking data. This will allow analysts to assess the off-ball contributions and movement of defenders with models such as pitch control \cite{spearman2018beyond}, evaluations of opposition marking and stopping events before they have happened. However, due to the low availability of tracking data at this time, especially across different leagues, it is key to have methods that allow defensive actions to be valued from event-data in smarter ways.

A further challenge our model faces is the dependency on another player metric. This becomes a problem in terms of interpretation, where people need to understand Expected Threat first in order to comprehend our DAxT model. Moreover, metrics such as Possession Adjusted Interceptions and True Tackle Win \% are slightly more intuitive than our ratings, which complicates the task for analytically less inclined scouts to fully grasp our model.

\section{Conclusion}\label{sect:conclusion}

Even though the results have illustrated that our model works well, comparing it to other baseline models remains difficult due to the lack of similar work. One analogous paper is \cite{stocklmaking}, where the authors make use of tracking data to understand the defensive impact of players.

In conclusion, we have presented a novel model for valuing defensive actions by using deep learning to predict a future event that a defensive action has stopped. We have introduced a new metric DAxT by focusing on the values of tackles and interceptions made by defenders. This can help clubs to better understand the contribution of defenders and identify new talent for recruitment in the over 100 leagues that event-data is collected for.

\bibliographystyle{ACM-Reference-Format}
\bibliography{references}

\clearpage

\appendix
\section{Appendix on Reproducibility}

In this appendix we give further details on our models and techniques used for reproducibility purposes.

\subsection{Data Conversion}

For this project, we used StatsBomb's event-based data for the 2017/2018 and 2018/2019 English Premier League seasons (total of 760 games). Due to the inconsistency of events in terms of number of features, we used a data abstraction package called SPADL (Soccer Player Action Description Language). As mentioned in \cite{decroos2019actions}, SPADL transforms the event stream data of JSON format into ``a common vocabulary that enables subsequent data analysis." This library was created to convert different data providers' event data into a consistent format. Therefore, our model is reproducible with any data. Table \ref{tab:SPADLexample} shows some features of the first 5 instances in the event-stream data.

\begin{table*}[bp]
    \caption{StatBomb's Premier League data converted to SPADL}
    \label{tab:SPADLexample}
    \small
    \centering
    \begin{tabular}{| c  c  c  c  c  c  c  c  c  c  c |}
    \toprule
    \textbf{Game ID} & \textbf{Time in seconds} & \textbf{Team ID} & \textbf{Player ID} & \textbf{Start x} & \textbf{Start y} & \textbf{...} & \textbf{Result ID} & \textbf{Type name} & \textbf{Player name} & \textbf{Country name}  \\
    \midrule
    14562  & 1 & 58 & 9923 & 52.0588 & 34.4304 & ... & 1 & pass  & Bobby Reid & England \\
    14562  & 2 & 58 & 9917 & 41.4706 & 34.4304 & ... & 1 & dribble  & Joe Ralls & England \\
    14562  & 2 & 58 & 9917 & 41.4706 & 34.4304 & ... & 1 & pass  & Joe Ralls & England \\
    14562  & 4 & 58 & 9924 & 25.5882 & 19.7975 & ... & 1 & dribble  & Sean Morrison & England \\
    14562  & 6 & 58 & 9924 & 26.4706 & 18.9367 & ... & 0 & pass  & Sean Morrison & England \\
    \bottomrule
    \end{tabular}
\end{table*}

\begin{figure}[h]
    \centering
    \includegraphics[width=\linewidth]{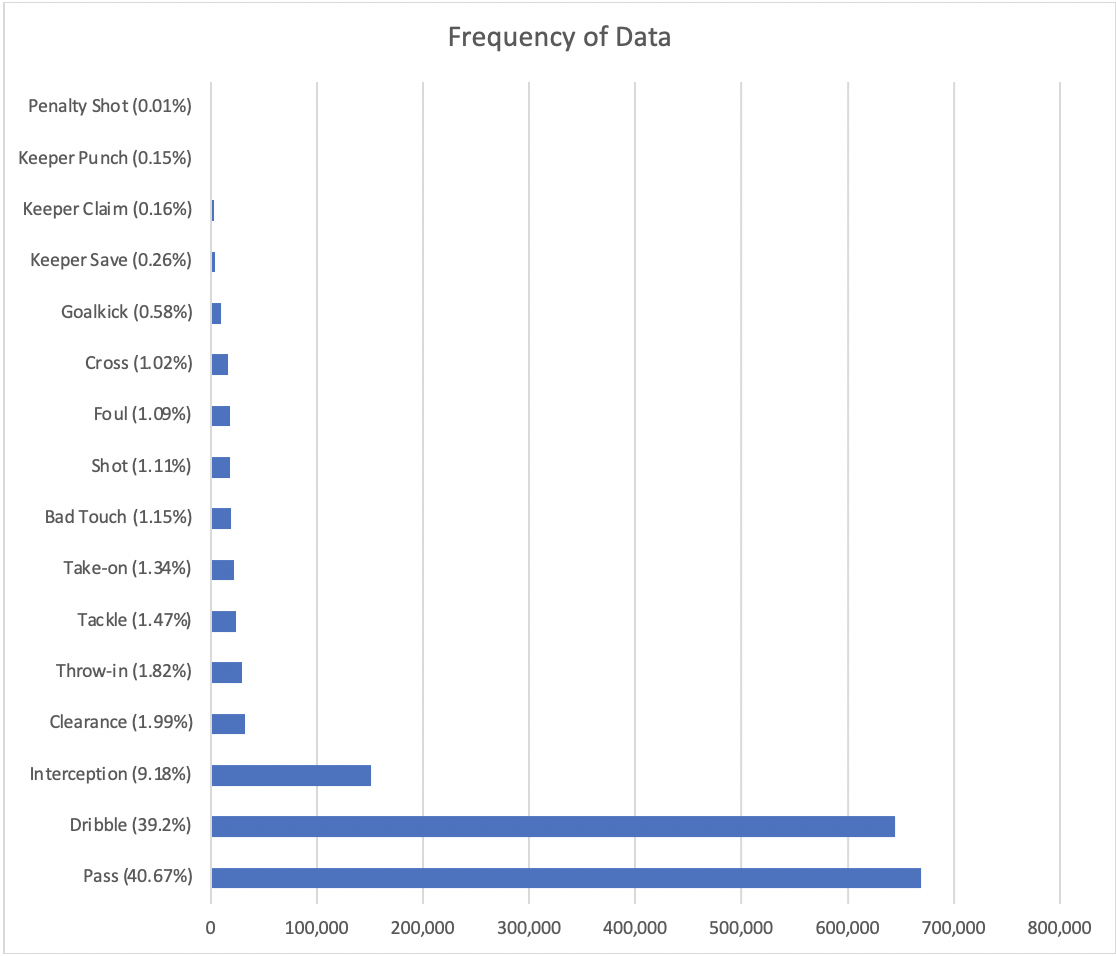}
    \caption{Frequency of each action type in StatBomb's data}
    \label{fig:freqStatbomb}
\end{figure}

We then calculated the xT of every moving successful action in the dataset. Figure \ref{fig:freqStatbomb} illustrates the frequency of each action type.

After preparing the data, we extracted all 3 consecutive, moving, and successful actions for our first database. Each extraction was 1 instance where 2 actions were considered as input (xT and coordinates) and the third one as output (xT only). This amounted to 802,046 instances. The next step was extracting similar sequences (2 actions) that happened before a failed event due to a defensive action by the opposition. We separated interceptions from tackles which resulted in 75,691 interceptions to value and 17,423 tackles.

Before training, we applied Scikit-Learn's MinMaxScaler \footnote{https://scikit-learn.org/} on our dataset for normalization.

\subsection{Model Setup}

For our model, we used the implementation of a Multi Layer Perceptron from Keras' library. The network was trained using a Mean Absolute Error loss function and an Exponential Linear Unit activation function. The deep network was composed of 3 dense hidden layers with 10 neurons each, and ran using Adadelta optimizer for 50 epochs.

For the experiments in the paper, we used the scipy.stats\footnote{https://docs.scipy.org/doc/scipy/reference/stats.html} library for the statistical outcomes and the matplotlib\footnote{https://matplotlib.org/} library for all graphs.

\end{document}